\PassOptionsToPackage{pagebackref}{hyperref}
\PassOptionsToPackage{hyphens}{url}
\documentclass[10pt,twocolumn,letterpaper]{article}

\usepackage{times}
\usepackage{soul}
\usepackage[utf8]{inputenc}
\usepackage[small]{caption}
\usepackage{graphicx}
\usepackage{amsmath}
\usepackage{amsthm}
\usepackage{booktabs}
\usepackage{algorithm}
\usepackage{algorithmic}
\usepackage{orcidlink}
\usepackage{multirow}
\usepackage{pifont}
\usepackage{amssymb}

\usepackage{epsfig}
\usepackage{color}
\usepackage{colortbl}
\usepackage{array}
\usepackage{textcomp}
\usepackage{stfloats}
\usepackage{verbatim}
\usepackage{graphicx}
\usepackage{tikz}
\usepackage{comment}
\usepackage{subcaption}
\usepackage[dvipsnames, svgnames, x11names]{xcolor}

\def\ie{{\em i.e.}}

\definecolor{mygray}{gray}{.9}
\definecolor{baselinecolor}{gray}{.9}
\newcommand{\baseline}[1]{\cellcolor{baselinecolor}{#1}}

\definecolor{myred}{RGB}{220, 50, 47}

\usepackage{wacv}              

\definecolor{wacvblue}{rgb}{0.21,0.49,0.74}

\def\wacvPaperID{362} 
\def\confName{WACV}
\def\confYear{2026}

\title{Structured Context Learning for Generic Event Boundary Detection}

\author{
    Xin Gu$^{1,3}$\thanks{Equal contributions \;\;\; $^{\dagger}$Corresponding author \;\;\; $\ddagger$This work was done during the first author’s internship at ByteDance} \;\;\;
    Congcong Li$^{2 *}$ \;\;\;
    Xinyao Wang$^3$ \;\;\;
    Dexiang Hong$^3$ \;\;\;
    Libo Zhang$^4$ \;\;\; \\
    Tiejian Luo$^{1\dagger}$ \;\;\;
    Longyin Wen$^3$ \;\;\;
    Heng Fan$^5$\\
    $^1$ University of Chinese Academy of Sciences
    $^2$ Sichuan Changhong Electric Co., Ltd. \\
    $^3$ByteDance Intelligent Creation
    $^4$Institute of Software, Chinese Academy of Sciences \\
    $^5$Department of Computer Science and Engineering, University of North Texas\\
}

\begin{document}

\maketitle

\begin{abstract}
Generic Event Boundary Detection (GEBD) aims to identify moments in videos that humans perceive as event boundaries. This paper proposes a novel method for addressing this task, called Structured Context Learning, which introduces the Structured Partition of Sequence (SPoS) to provide a structured context for temporal information learning. Our approach is end-to-end trainable and flexible, not being restricted to specific temporal models like GRU, LSTM, and Transformers. This flexibility enables our method to achieve a better speed-accuracy trade-off. Specifically, we apply SPoS to partition the input frame sequence and provide a structured context for the subsequent temporal model. Notably, SPoS's overall computational complexity is linear with respect to the video length. We next calculate group similarities to capture differences between frames, and a lightweight fully convolutional network is utilized to determine the event boundaries based on the grouped similarity maps. To remedy the ambiguities of boundary annotations, we adapt the Gaussian kernel to preprocess the ground-truth event boundaries. Our proposed method has been extensively evaluated on the challenging Kinetics-GEBD, TAPOS, and shot transition detection datasets, demonstrating its superiority over existing state-of-the-art methods.
\end{abstract}

\section{Introduction}

Video has become an integral part of human life in recent years, and with rapid developments in hardware, video understanding has witnessed a surge in newly designed architectures \citep{DBLP:conf/icml/3D-Convolutional,DBLP:conf/iccv/Learning-Spatiotemporal,DBLP:conf/iccv/SlowFast,DBLP:conf/cvpr/Two-Stream,DBLP:journals/corr/abs-2102-00719,DBLP:journals/corr/abs-2106-13230} and datasets \citep{DBLP:journals/corr/Kinetics,DBLP:journals/corr/UCF101,DBLP:conf/cvpr/TAPOS,DBLP:conf/iccv/HMDB,DBLP:conf/cvpr/PerazziPMGGS16}. Cognitive science \citep{tversky2013event} suggests that humans naturally divide videos into meaningful units. To enable machines to develop this ability, Generic Event Boundary Detection (GEBD) \citep{shou2021generic} was recently proposed, which aims to localize moments where humans naturally perceive event boundaries. The GEBD task deals with taxonomy-free event boundaries and attempts to connect human perception mechanisms with video understanding. Annotators are required to localize boundaries at a more granular level compared to video-level events. To address the ambiguities of event boundaries based on human perception, five different annotators label each video's boundaries based on predefined principles. These characteristics differentiate GEBD from previous video localization tasks \citep{DBLP:journals/access/XiaZ20}. The GEBD task is more challenging due to several high-level factors, including changes in subject, action, environment, and object of interaction. These factors make GEBD a more demanding task compared to video localization. Solving GEBD is a challenging task, as it heavily relies on temporal context information. Existing methods address this issue by processing each frame individually \citep{shou2021generic,DBLP:conf/cvpr/DMC-Net,DBLP:journals/corr/abs-2107-00239}, or by computing the global self-similarity matrix and utilizing extra parsing algorithms to identify boundary patterns \citep{DBLP:journals/corr/gebd-UBoCo,DBLP:journals/corr/gebd-contrastive-learning}. However, the former methods suffer from significant redundancy in computations of adjacent frames in a video sequence and the class imbalance issue of event boundaries. The latter methods have quadratic computational complexity with respect to the length of input videos, due to the computation of self-attention globally.
We propose an efficient end-to-end method for predicting all boundaries of video sequences in a single forward pass of the network. The Structured Context Learning (SCL) model is designed for GEBD based on the Structured Partition of Sequence (SPoS) mechanism. This mechanism has linear computational complexity with respect to the length of the input video and enables feature sharing. The SPoS mechanism brings the local feature sequences for each frame in a one-to-one manner, which is termed the \textbf{structured context}. The design of SPoS also allows for flexibility in choosing different temporal models without losing much performance, achieving a better speed-accuracy trade-off. We observe that 1D CNNs are suboptimal for boundary detection as adjacent frames are equally important, while 1D CNNs actually make the candidate frames attend to adjacent frames in a Gaussian distribution manner \citep{DBLP:conf/nips/LuoLUZ16}. Our proposed SCL model can learn a high-level representation for each frame within its structured context, which is critical for boundary detection. We then utilize group similarity learning to enable the network to learn a diverse set of similarities, which is highly effective for GEBD. Following the group similarity, we use a lightweight fully convolutional network (FCN)~\citep{DBLP:conf/cvpr/FCN} to predict the event boundaries. To speed up training, we use a Gaussian kernel to preprocess the ground-truth event boundaries. We conduct extensive experiments on two challenging datasets, Kinetics-GEBD and TAPOS, which demonstrate the effectiveness of our method compared to state-of-the-art methods. On the Kinetics-GEBD dataset, our proposed method achieves a 2.0\% absolute improvement over DDM-Net while running 8.6$\times$ faster, and a 15.9\% absolute improvement over PC~\citep{shou2021generic} while running 5.7$\times$ faster. We also evaluated shot transition detection datasets and conducted several ablation studies to analyze the effectiveness of different components in our proposed method. 

In summary, our main contributions are as follows: \ding{171} We propose a structured context learning method for GEBD, which can be trained in an end-to-end fashion; \ding{170} We propose Structured Partition of Sequence (SPoS) to address context learning in GEBD,  which has linear computation complexity and is flexible in choosing different temporal models; \ding{168} We compute group similarities to exploit discriminative features and encode the differences between frames; \ding{169} Experiments on two challenging Kinetics-GEBD and TAPOS datasets demonstrate the effectiveness of our method compared to SOTA methods. Additionally, we apply our method to the shot detection datasets ClipShots, BBC and RAI, achieving superior results and demonstrating its potential for similar tasks.

\section{Related Works}

\textbf{Temporal Action Localization (TAL).} TAL aims to precisely locate  action segments in untrimmed videos by detecting their start and end points along with their corresponding action class. Two distinct approaches have been developed to address this challenge, including two-stage \citep{caba2017scc,zhao2017temporal,Chao_2018_CVPR} and single-stage methods \citep{alwassel2018action,long2019gaussian,zhao2020bottom}. In two-stage methods, the first stage generates proposals of action segments, while the second stage determines the actionness and type of action for each proposal. Additionally, grouping \citep{zhao2017temporal} and Non-maximum Suppression (NMS) \citep{DBLP:conf/iccv/BMN} are used to eliminate redundant proposals. In contrast, the single-stage method performs classification on pre-defined anchors \citep{lin2017single,long2019gaussian} or video frames \citep{Yuan_2017_CVPR}. Although the TAL and GEBD tasks share some similarities, the latter requires event boundaries to be both taxonomy-free and continuous, which differs from the settings in TAL. Thus, direct application of TAL methods on GEBD is not straightforward.

\vspace{0.3em}
\noindent
\textbf{Generic Event Boundary Detection (GEBD).} The objective of GEBD~\citep{shou2021generic} is to identify event boundaries that divide a long event into shorter, continuous segments without relying on any predefined taxonomy. Unlike TAL, GEBD focuses only on predicting segment boundaries. Current GEBD methods~\citep{DBLP:journals/corr/gebd-contrastive-learning,DBLP:journals/corr/abs-2107-00239,DBLP:journals/corr/abs-2106-10090} adopt a similar approach to \citep{shou2021generic}. They take a fixed number of video frames before and after a candidate frame as input and then determine whether that frame is an event boundary or not. Kang \textit{et al.} \citep{DBLP:journals/corr/gebd-contrastive-learning} propose using the temporal self-similarity matrix (TSM) as an intermediate representation and then use contrastive learning to extract discriminative features for improved performance. Hong \textit{et al.} \citep{DBLP:journals/corr/abs-2107-00239} use a cascade classification approach and a dynamic sampling strategy to increase both recall and precision. Rai \textit{et al.} \citep{DBLP:journals/corr/abs-2106-10090} learn spatiotemporal features using a two-stream inflated 3D convolutional architecture.

\vspace{0.3em}
\noindent
\textbf{Context Learning in Sequences.} Context is the information surrounding a sequence of data, which is crucial in tasks such as natural language processing, speech recognition, and image recognition, where the interpretation of an element in the sequence is influenced by its neighboring elements. Recurrent neural networks, including LSTM \citep{DBLP:journals/neco/lstm} and GRU\citep{DBLP:journals/corr/GRU}, as well as Transformer-based models, are widely used to capture context in deep learning models. While Transformers have achieved remarkable results in various fields, the self-attention mechanism's computational complexity is $O(n^2)$ relative to the input sequence length, making it unsuitable for lengthy sequence data. To address this issue, different methods \citep{DBLP:conf/iccv/swin-transformer,DBLP:journals/corr/Longformer,DBLP:conf/cvpr/TCANet,DBLP:journals/corr/FaceTransformer,DBLP:conf/iccv/T2TViT} have been proposed to reduce computational burden by adapting local sliding window mechanisms, which have linear computational complexity with respect to image size. However, we found that these approaches are not suitable for GEBD tasks since stacked sliding windows tend to fuse different boundaries, and distant frames can cross multiple boundaries, leading to convergence disruptions. In this paper, we propose a novel structured context learning method for GEBD tasks that can be extended to similar tasks.

\section{Method}
\label{sec:method}

\begin{figure*}[t]
    \centering
    \includegraphics[width=0.92\linewidth]{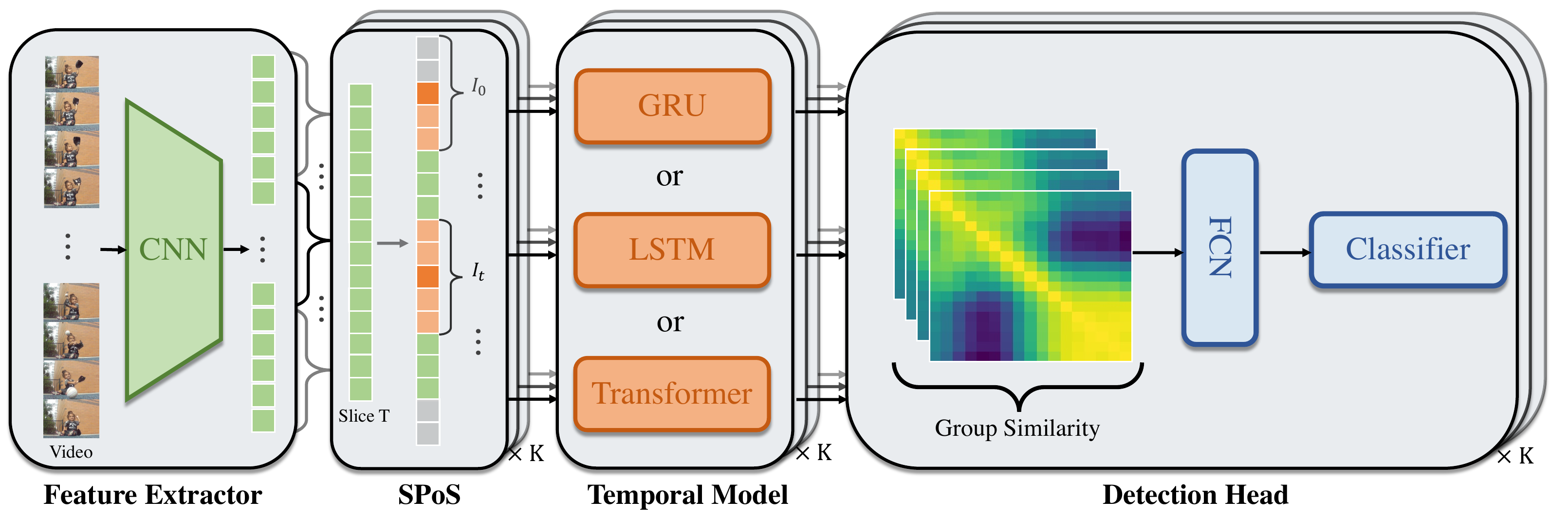}
    \caption{Proposed Structured Context Learning method. A CNN backbone extracts 2D features that are pooled and converted to a sequence. The SPoS mechanism partitions the sequence, providing structured context $\mathbf{I}_t$. Temporal model (Transformer LSTM, or GRU) learns high-level representations and enable feature sharing. Group similarities encode frame differences, and a lightweight FCN predicts event boundaries based on 2D grouped similarity maps.}
    \label{fig:framework}
    \vspace{-0.5em}
\end{figure*}

Previous methods \citep{shou2021generic,DBLP:conf/cvpr/DMC-Net,MA-net} approach the GEBD task as binary classification, predicting the boundary label of each frame by considering the temporal context. However, this method is inefficient as it requires redundant computation when generating consecutive frame representations. To address this issue, we propose a straightforward, end-to-end efficient method that considers each video clip as a whole. Firstly, we extract the feature representation for each frame using a conventional CNN backbone. This results in a frame sequence $V = \{ I_t\}_{t=1}^T$, where $I_t \in \mathbb{R}^C$ and $T$ is the length of the video clip. We then utilize the Structured Partition of Sequence (SPoS) mechanism to re-partition the input frame sequence $\{ I_t\}_{t=1}^T$ and provide \textbf{structured context} for each candidate frame. A temporal model is then used to learn the high-level representation of each local sequence. We capture temporal changes by computing group similarities and then use a lightweight fully convolutional network \citep{DBLP:conf/cvpr/FCN} (FCN) to recognize various patterns in the grouped 2D similarity maps. We will provide further details about each module in the subsequent sections. The overall architecture of our proposed method is depicted in Figure \ref{fig:framework}.

\subsection{Structured Context Learning}
\label{sec:sc_learning}

The presence of an event boundary in a video clip indicates a change in visual content at that particular point, making it challenging to determine the boundary from a single frame. As a result, localizing changes in the temporal domain is crucial for event boundary detection. Several approaches, including RNN, Transformer \citep{DBLP:conf/nips/att_is_all_you_need}, and 3D CNN\citep{DBLP:conf/iccv/Learning-Spatiotemporal}, have explored modeling in the temporal domain. While the Transformer has shown promising results in natural language processing and computer vision tasks, applying it directly to the GEBD task is challenging due to its quadratic computational complexity of self-attention, which results in a significant increase in computation costs and memory consumption as the length of the video increases. Previous methods \citep{shou2021generic,DBLP:conf/cvpr/DMC-Net} consider each frame as an individual sample, and its neighboring frames are fed into the network together to provide temporal information. However, this approach introduces redundant computation in adjacent frames, as each frame is fed into the network as input multiple times. In this paper, we aim to explore a more efficient and general temporal representation for the GEBD task.

\vspace{0.3em}
\noindent{\textbf{Structured Partition of Sequence:}} The objective is to obtain a sequence of $K$ adjacent frames before and after a given frame $I_t$, where $K$ is the adjacent window size, from a video snippet $V = \{ I_t\}_{t=1}^T$. Here, $T$ is the time span of the video snippet, and $I_t \in \mathbb{R}^C$ is the feature vector of frame $t$ generated from a ResNet50 \citep{DBLP:conf/cvpr/resnet} backbone, followed by a global average pooling layer. We refer to this local sequence centered on frame $I_t$ as \textbf{structured context}. To achieve this while maintaining feature sharing, efficiency, and parallelism, we introduce the novel Structured Partition of Sequences (SPoS) mechanism. Specifically, we pad the video $V = \{ I_t\}_{t=1}^T$ with $\text{ceil}(\frac{T}{K}) \cdot K - T$ zero vectors at the end of the frame sequence to ensure that the new video length $T'$ is divisible by $K$. Next, we split the padded video $V' \in \mathbb{R}^{T' \times C}$ into $K$ slices, where each slice $S_k$ (with slice number $k$ starting from $0$) provides structured context frames for all $[k::K]$th frames (i.e., all frames that start from $k$ with a step of $K$). Thus, all video frames can be covered within these $K$ slices, which can be efficiently processed in parallel.

\begin{figure*}[t]
    \centering
    \includegraphics[width=0.92\textwidth]{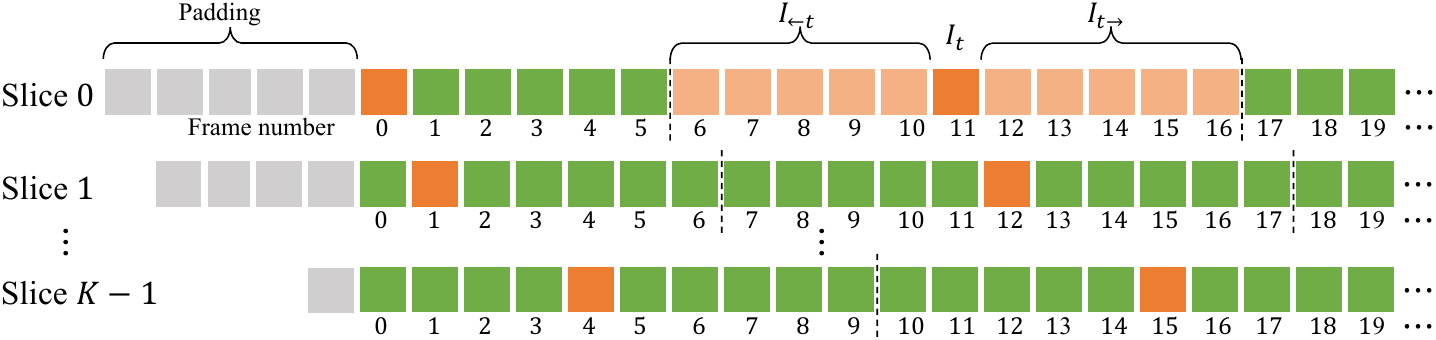}
    \caption{Illustration of proposed SPoS. The dark orange square denotes the candidate frame $I_t$, while the light orange squares denote its structured context. To obtain adjacent $K$ frames $I_{\leftarrow t}$ before candidate frame $I_t$ and $K$ frames $I_{t \rightarrow}$ after $I_t$ , we split the input video sequence into $K$ slices. Each slice $S_k$ is responsible to produce adjacent frames $I_{\leftarrow t}$ and $I_{t \rightarrow}$ for the frames of specific indices. All video frames can be covered within all $K$ slices and can be efficiently processed in parallel.}
    \label{fig:spos}
    \vspace{-0.3em}
\end{figure*}

In each frame slice $S_k$, we obtain a structured context for frame $I_t$ in two directions: $K$ frames before and $K$ frames after. We implement this efficiently using the memory view method provided by modern deep learning frameworks. Specifically, to obtain the structured context frames $I_{\leftarrow t} \in \mathbb{R}^{K \times C}$ before frame $I_t$, we replicate the first frame of the padded video sequence $V'$ $K-k$ times, concatenate it to the beginning of video $V'$, and drop the last $K-k$ frames to keep the number of frames divisible by $K$. This yields a shifted video sequence denoted as $V'_\leftarrow \in \mathbb{R}^{T' \times C}$, which we then view as $V_\leftarrow \in \mathbb{R}^{N \times K \times C}$, where $N = \frac{T'}{K}$ denotes the number of processed frames in the slice $S_k$. This gives us the left structured context frames for all $N$ frames, i.e., all $[k::K]$th frames of the original video $V$. Similarly, to obtain the structured context frames $I_{t\rightarrow} \in \mathbb{R}^{K \times C}$ after frame $I_t$, we replicate the last frame of the padded video sequence $k+1$ times, concatenate it to the end of video $V'$, and drop the first $k+1$ frames. This gives us the right structured context frames $V_\rightarrow \in \mathbb{R}^{N \times K \times C}$ for all $N$ frames. Finally, we obtain all temporal context frames by repeating this process $K$ times for $K$ slices. Each frame $I_t$ is represented by its adjacent frames in a local window. Figure~\ref{fig:spos} illustrates SPoS.

SPoS is a critical design element of our approach, thanks to its shared structured context information. We refer to this context information as ``structured'' because SPoS maps each candidate frame ($I_t$) to individual frame sequences ($I_{\leftarrow t}$ and $I_{t\rightarrow}$) in a one-to-one manner, which is essential for accurate boundary detection. In contrast, other sliding window based methods \citep{DBLP:conf/iccv/swin-transformer,DBLP:journals/corr/Longformer,DBLP:conf/cvpr/TCANet,DBLP:journals/corr/FaceTransformer,DBLP:conf/iccv/T2TViT} enable each frame to attend to very distant frames by learning a global representation due to the stacked window design. However, this approach is detrimental to boundary detection, as very distant frames may span multiple boundaries and contribute less relevant information. An additional advantage of using SPoS is that it enables structured sequence modeling with any sequential method, without concerns about computational complexity. Thanks to its local, shared, and parallel design, SPoS can be computed in linear time with respect to video length.

\vspace{0.3em}
\noindent{\textbf{Encoding with Temporal Model:}} SPoS's flexible design allows us to use any temporal model to learn semantic features. In this work, we use the Transformer as an example to demonstrate how to obtain structured context information. Given structured context features $I_{\leftarrow t} \in \mathbb{R}^{K \times C}$ and $I_{t\rightarrow} \in \mathbb{R}^{K \times C}$ of frame $I_t \in \mathbb{R}^C$, we concatenate them in the temporal dimension to obtain the context sequence $\mathbf{I}_t$ for frame $I_t$ as follows:
\begin{equation}
    \mathbf{I}_t = [I_{\leftarrow t}, I_t, I_{t\rightarrow}]
\end{equation}
where $\mathbf{I}_t \in \mathbb{R}^{L \times C}$, $L=2K+1$, and $[\cdot, \cdot, \cdot]$ denote the concatenation operation. To model the temporal information, we adapt a 6-layer standard Transformer \citep{DBLP:conf/nips/att_is_all_you_need} block to process the context sequence $\mathbf{I}_t$ and obtain the temporal representation $\mathbf{x}_t \in \mathbb{R}^{L \times C}$ within the structured context window. Unlike other methods \citep{DBLP:journals/corr/gebd-contrastive-learning,DBLP:journals/corr/gebd-UBoCo}, where the computation of multi-head self-attention is based on the global video frame sequence, our computation is based only on the local temporal window. The computational complexity of the former is quadratic relative to the video length $T$, \ie, $4TC^2 + 2T^2C$, while the computational complexity of our method is linear when $K$ is fixed, \ie, $4TC^2 + 2L^2TC$. Global self-attention computation is generally unaffordable for large video lengths $T$, whereas our local structured-based self-attention is scalable.

\subsection{Group Similarity}

The event boundaries in the GEBD task can be identified when there is a change in action, subject, or environment. In our experiments, we observed that adjacent frames within a local window provided more useful information for detecting event boundaries than distant frames. This is consistent with human intuition, as changes in visual content can be considered event boundaries only in short time periods. Based on this observation, we can naturally model local temporal information using the structured context features that we extracted in Section \ref{sec:sc_learning}.

\begin{figure}
  \centering
  \subfloat{\includegraphics[width=0.22\linewidth]{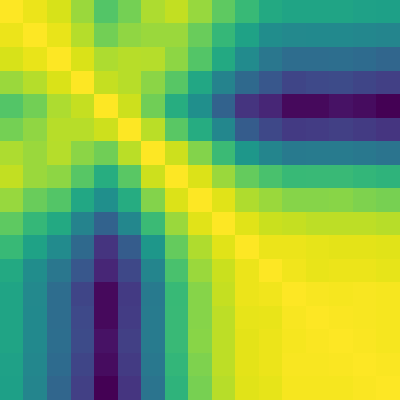}}\hspace{1pt}
  \subfloat{\includegraphics[width=0.22\linewidth]{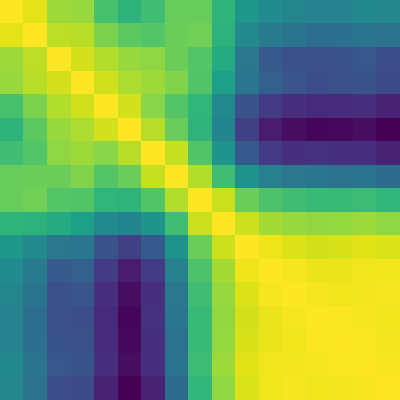}}\hspace{1pt}
    \subfloat{\includegraphics[width=0.22\linewidth]{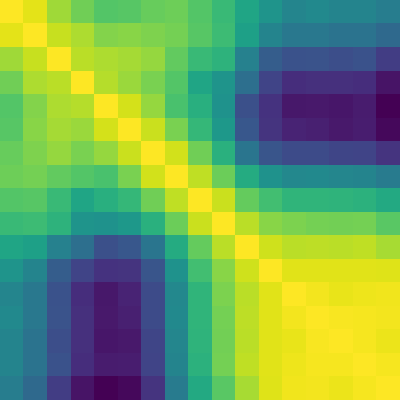}}\hspace{1pt}
  \subfloat{\includegraphics[width=0.22\linewidth]{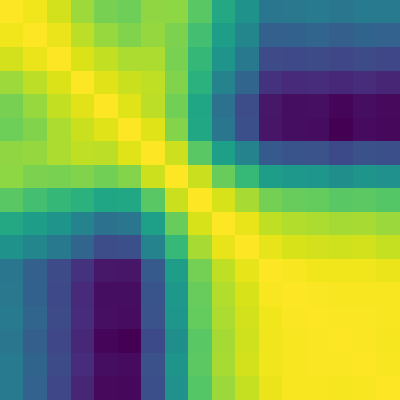}}\\ [0.1in]
   \subfloat{\includegraphics[width=0.22\linewidth]{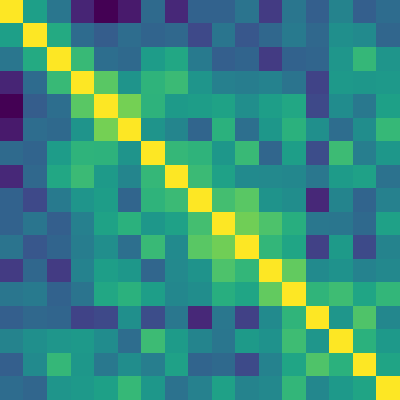}}\hspace{1pt}
  \subfloat{\includegraphics[width=0.22\linewidth]{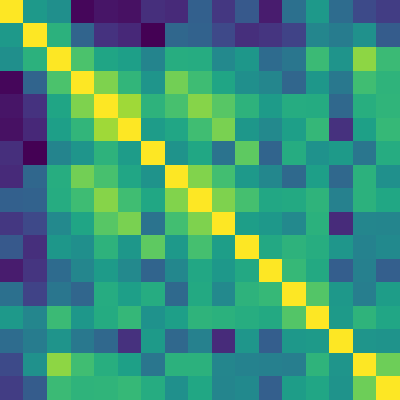}}\hspace{1pt}
    \subfloat{\includegraphics[width=0.22\linewidth]{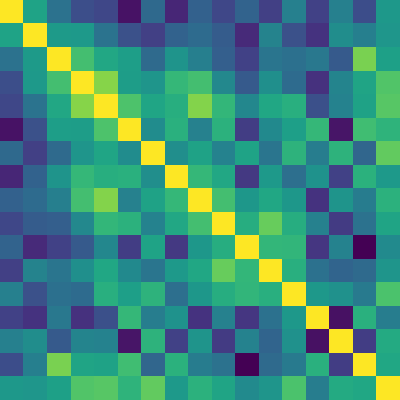}}\hspace{1pt}
  \subfloat{\includegraphics[width=0.22\linewidth]{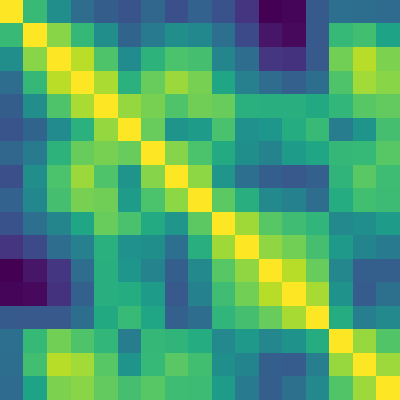}}
  \caption{Visualization of grouped similarity maps $\mathbf{S}_t$, $G=4$ in this example. First row indicates that there is a potential boundary in this local sequence while the second row shows no boundary in this sequence. We can also observe slightly different patterns between the same group, which may imply that each group is learning in a different aspect.}
\label{fig:similarity_maps}
\end{figure}

The event boundaries highlight differences between adjacent frames, and neural networks can take shortcuts during learning \citep{DBLP:journals/natmi/shortcut-learning}. Therefore, classifying frames directly as boundaries may yield inferior performance due to the lack of explicit cues. To address this, we propose guiding classification with feature similarity between each frame pair in the structured temporal window $\mathbf{x}_t \in \mathbb{R}^{L \times C}$. We divided the $C$-dimensional channels into several groups and calculate the similarity of each group independently. The concept of grouping as a dimension for model design has been widely studied, including Group Convolutions \citep{DBLP:conf/nips/alexnet,DBLP:conf/cvpr/ResNeXt}, Group Normalization \citep{DBLP:conf/eccv/GN}, and Multi-Head Self Attention \citep{DBLP:conf/nips/att_is_all_you_need}. However, as far as we know, there is no study on similarity learning with grouping. Given $\mathbf{x}_t \in \mathbb{R}^{L \times C}$, we first split it into $G$ groups:
\begin{equation}
    \mathbf{x'}_t = \text{reshape}(\mathbf{x}_t)
\end{equation}
where $\mathbf{x'}_t \in \mathbb{R}^{L \times G \times C'}$ and $C' = \frac{C}{G}$. Then the group similarity map $\mathbf{S}_t$ is calculated using the grouped feature:
\begin{equation}
\label{equa:sim_func}
    \mathbf{S}_t = \text{similarity-function}(\mathbf{x'}_t,\mathbf{x'}_t)
\end{equation}
where $\mathbf{S}_t \in \mathbb{R}^{G \times L \times L}$, and $\text{similarity-function}(\cdot, \cdot)$ can be any similarity metric. The group similarity map $\mathbf{S}_t$ contains efficient scores for each pair of frames, indicating high response values when the frames are visually similar. These similarity patterns are critical for detecting event boundaries and vary between different sequences, as demonstrated in Figure \ref{fig:similarity_maps}. To maintain simplicity in our model, we adapt a 4-layer fully convolutional network \citep{DBLP:conf/cvpr/FCN} to learn these patterns, which we find to be both effective and efficient. We then perform average pooling on the output of the FCN to obtain a vector representation $h_t$:
\begin{equation}
    h_t = \text{average-pool}(\operatorname{FCN}(\mathbf{S}_t))
\end{equation}
where $h_t \in \mathbb{R}^C$. The design principle of this module is straightforward and intuitive: it computes group similarity patterns within a local structured context based on the previously encoded frames and uses a small FCN to analyze the patterns.

\subsection{Optimization}
\label{sec:optim}

Our lightweight and fully end-to-end Structured Context Learning model, combined with group similarity, facilitates direct use for further classification. This simplicity makes implementation and optimization a straightforward process. The video frame sequence $V = \{ I_t\}_{t=1}^T$ is represented by $\mathbf{V} = \{h_t\}_{t=1}^T$ after the group similarity module, resulting in an output $\mathbf{V} \in \mathbb{R}^{T \times C}$ with no dimensional change between input and output. We then stack 3 layers of 1D convolutional neural networks to predict boundary scores. To optimize our network, we use a single binary cross-entropy loss.

GEBD is a taxonomy-free task that connects human perception to deep video understanding. Since around 5 annotators annotate each video's event boundary labels to capture differences in human perception, diversity is ensured. However, this can also lead to annotation ambiguity and poor convergence during network optimization. To address this, we use a Gaussian kernel to smooth the ground-truth boundary labels and obtain soft labels instead of hard labels. Specifically, for each annotated boundary, we compute the intermediate label of the neighboring position $t'$ by utilizing the Gaussian distribution:
\begin{equation}
    \mathcal{L}_{t'}^t = \exp\Big( -\frac{( t-t' )^2}{2\sigma^2} \Big)
\end{equation}
where $\mathcal{L}_{t'}^t$ indicates the intermediate label at time $t'$ corresponding to the annotated boundaries at time $t$. We set $\sigma =1$ in all our experiments. The final soft labels are obtained by summing all intermediate labels. The binary cross-entropy loss is then used to minimize the difference between the model predictions and the soft labels.

\section{Experiments}

Our method achieves competitive results compared to previous approaches in Kinetics-GEBD \citep{shou2021generic} and TAPOS \citep{DBLP:conf/cvpr/TAPOS} datasets, as shown in our quantitative evaluation. We further demonstrate the potential of our approach by evaluating shot transition detection datasets, including ClipShots~\citep{DBLP:conf/accv/TangFKCZ18}, BBC~\citep{DBLP:conf/mm/BBC_Dataset}, and RAI~\citep{DBLP:conf/caip/RAI_Dataset}. Additionally, we conduct a detailed ablation study of our model design, providing insights and quantitative results on the GEBD task.

\subsection{Dataset}
We first evaluate our proposed method on two datasets: Kinetics-GEBD \citep{shou2021generic} and TAPOS \citep{DBLP:conf/cvpr/TAPOS}. The Kinetics-GEBD dataset consists of $54,691$ videos and $1,290,000$ temporal boundaries spanning various video domains in the wild. We train and test our model on the training and validation sets, respectively, as the ground truth labels for the testing videos have not been released. The TAPOS dataset comprises Olympic sport videos in 21 action classes, with $13,094$ and $1,790$ action instances in the training and validation sets, respectively. We re-purpose the TAPOS dataset for the GEBD task following \citep{shou2021generic} by trimming each action instance with its action label hidden. In addition, we evaluate our method on shot transition detection datasets: ClipShots \citep{DBLP:conf/accv/TangFKCZ18}, BBC \citep{DBLP:conf/mm/BBC_Dataset}, and RAI \citep{DBLP:conf/caip/RAI_Dataset}. ClipShots contains 128,636 cut transitions and 38,120 gradual transitions from 4,039 online videos, BBC has around 4,900 shots and 670 scenes, and RAI contains 987 shot boundaries.

\subsection{Evaluation Protocol}
We use the F1 score as the quantitative evaluation metric for GEBD task. To determine whether a detection is correct or not, we use the Rel.Dis. (Relative Distance) metric as described in \citep{shou2021generic}, which calculates the error between the detected and ground truth timestamps divided by the length of the corresponding action instance. The highest F1 score is taken as the final result after comparing the detection against each rater's annotation. We report F1 scores for different threshold values ranging from 0.05 to 0.5 with a step of 0.05. For ClipShots, BBC and RAI datasets, we use the evaluation standard in TransNet V2 \citep{DBLP:journals/corr/abs-2008-04838} and report the corresponding F1 scores.

\subsection{Implementation Details}

To ensure a fair comparison with other methods, we use a ResNet50 \citep{DBLP:conf/cvpr/resnet} pretrained on ImageNet \citep{DBLP:conf/cvpr/imagenet} as the basic feature extractor in all experiments unless otherwise specified. We optimize the ResNet50 parameters through backpropagation without freezing them. Following \citep{shou2021generic}, we resize images to 224$\times$224. We sample 100 frames uniformly from each video for batching, which corresponds to $T=100$ in Section \ref{sec:method}. We use the standard SGD with a momentum of $0.9$, a weight decay of $10^{-4}$, and a learning rate of $10^{-2}$. Each GPUs is assigned a batch size of $4$ (i.e., 4 videos, equivalent to 400 frames), and we train the network on $8$ NVIDIA Tesla V100 GPUs, resulting in a total batch size of $32$. We use automatic mixed precision training to reduce the memory burden. The network is trained for $30$ epochs, with the learning rate decreasing by a factor of $10$ after $16$ and $24$ epochs, respectively.

\begin{table*}[!t]
\centering
\renewcommand{\arraystretch}{1.2}
\resizebox{0.91\linewidth}{!}{
\begin{tabular}{lcccccccccc|c}
\rowcolor{mygray}
\specialrule{1.5pt}{0pt}{0pt}
Rel.Dis. threshold& 0.05 & 0.1 & 0.15 & 0.2 & 0.25 & 0.3 & 0.35 & 0.4 & 0.45 & 0.5 & \textbf{Avg.} \\ 
\hline
PC \textcolor{lightgray}{\scriptsize{[ICCV'21]}}~\citep{shou2021generic}& 0.625& 0.758&0.804&0.829&0.844&0.853&0.859&0.864&0.867&0.870&0.817\\
PC + OF \textcolor{lightgray}{\scriptsize{[ICCV'21]}}~\citep{li2022end}& 0.646& 0.776&0.818&0.842&0.856&0.864&0.868&0.874&0.877&0.879&0.830\\
UBoCo \textcolor{lightgray}{\scriptsize{[CVPR'22]}}~\citep{kang2022uboco} &0.702&0.846&0.862&0.879&0.888&0.889&0.895&0.897&0.904&0.905&0.866 \\
Com-Net \textcolor{lightgray}{\scriptsize{[CVPR'22]}}~\citep{li2022end} &0.743&0.830&0.857&0.872&0.880&0.886&0.890&0.893&0.896&0.898&0.865 \\
DDM-Net \textcolor{lightgray}{\scriptsize{[CVPR'22]}}~\citep{DBLP:conf/cvpr/DMC-Net}&0.764&0.843&0.866&0.880&0.887&0.892&0.895&0.898&0.900&0.902&0.873\\
TP-Net \textcolor{lightgray}{\scriptsize{[TPAMI'23]}}~\citep{tan2023temporal} &0.748&0.828&0.852&0.866&0.874&0.879&0.883&0.887&0.890&0.892&0.860 \\
MA-Net \textcolor{lightgray}{\scriptsize{[WACV'23]}}~\citep{MA-net}&0.680&0.779&0.806&0.818&0.825&0.830&0.834&0.837&0.839&0.841&0.809\\
BasicGEBD \textcolor{lightgray}{\scriptsize{[ACM MM'24]}}~\citep{zheng2024rethinking}&0.768&0.834&-&-&-&0.885&-&-&-&0.896&0.866\\
\midrule
Ours& \textbf{0.784} & \textbf{0.856} & \textbf{0.877} & \textbf{0.890} & \textbf{0.896} & \textbf{0.901} & \textbf{0.904} & \textbf{0.907} & \textbf{0.909} & \textbf{0.911} & \textbf{0.883} \\
\specialrule{1.5pt}{0pt}{0pt}
\end{tabular}
}
\caption{F1 results on Kinetics-GEBD validation split with Rel.Dis. threshold set from 0.05 to 0.5 with 0.05 interval.}
\label{tab:gebd_val}
\end{table*}

\subsection{Main Results}

\noindent{\textbf{Kinetics-GEBD.}}
Table \ref{tab:gebd_val} presents the performance of our models on the Kinetics-GEBD validation set. Our method outperforms all previous methods in all Rel.Dis. threshold settings demonstrate the effectiveness of the Structured Context Learning method. Compared to the PC method, our method achieves a significant 15.9\% absolute improvement with a $5.7\times$ faster running speed (\ie, 10.8ms / frame vs 1.9ms / frame). In addition, compared to the SOTA method DDM-Net, we also achieve a 2.0\% absolute improvement with an $8.6\times$ faster running speed (\ie, 16.3ms / frame vs 1.9ms / frame), based on the official implementation. Both DDM-Net and PC employ the same input representation, in which each frame and its adjacent frames are fed into the network individually, resulting in many redundant computations. SPoS enables shared, parallel computation across all windows, reducing redundancy and improving context modeling, which consistently leads to significant performance gains.

\noindent{\textbf{TAPOS.}} The TAPOS dataset~\citep{DBLP:conf/cvpr/TAPOS} contains Olympic sport videos featuring 21 actions. Although not originally intended for the GEBD task, we repurposed the dataset by trimming each action instance with its action label hidden following \citep{shou2021generic}. This results in a more fine-grained sub-action boundary detection dataset. The results are presented in Table \ref{tab:tapos_val}. Our method achieves a 10.4\% improvement compared to PC and a 1.8\% improvement compared to the state-of-the-art method TP-Net. These results demonstrate the effectiveness of our method and its ability to learn more robust feature representations in various scenes.

\noindent{\textbf{ClipShots, BBC, RAI.}}
In addition to evaluating our approach on the Kinetics-GEBD and TAPOS datasets, we also tested its performance on shot transition detection datasets, namely ClipShots~\citep{DBLP:conf/accv/TangFKCZ18}, BBC~\citep{DBLP:conf/mm/BBC_Dataset}, and RAI~\citep{DBLP:conf/caip/RAI_Dataset}. We used ShuffleNet v2 0.5x \citep{DBLP:conf/eccv/ShuffleNetV2} and GRU in these experiments. These datasets include both hard cuts and gradual changes, which can result in false hits and false dismissals, making it essential for the model to learn the semantics of the video and distinguish between real transitions. Our results, presented in Table~\ref{tab:shot_results}, show that our approach achieves significant improvements in the F1 score, with an increase of 1. 6\%, 2. 2\%, and 2. 3\% in the ClipShots, BBC, and RAI datasets, respectively. These cross-dataset evaluation results demonstrate the strong generalization and robustness of our method.

\begin{table*}[!t]
\centering
\renewcommand{\arraystretch}{1.2}
\resizebox{0.95\linewidth}{!}{
\begin{tabular}{lcccccccccc|c}
\rowcolor{mygray}
\specialrule{1.5pt}{0pt}{0pt}
Rel.Dis. threshold& 0.05 & 0.1 & 0.15 & 0.2 & 0.25 & 0.3 & 0.35 & 0.4 & 0.45 & 0.5 & \textbf{Avg.} \\
\hline
TransParser \textcolor{lightgray}{\scriptsize{[CVPR'20]}}~\citep{DBLP:conf/cvpr/TAPOS}& 0.289 & 0.381 & 0.435 & 0.475 & 0.500 & 0.514 & 0.527 & 0.534 & 0.540&0.545&0.474 \\
PA \textcolor{lightgray}{\scriptsize{[ICCV'21]}}~\citep{shou2021generic}& 0.360 & 0.459 & 0.507 & 0.543 & 0.567 & 0.579 & 0.592 & 0.601 & 0.609 &0.615 &0.543 \\
PC \textcolor{lightgray}{\scriptsize{[ICCV'21]}}~\citep{shou2021generic}& 0.522 & 0.595 & 0.628 & 0.646 & 0.659 & 0.665 & 0.671 & 0.676 & 0.679 & 0.683 & 0.642 \\
DDM-Net \textcolor{lightgray}{\scriptsize{[CVPR'22]}}~\citep{DBLP:conf/cvpr/DMC-Net}& 0.604 & 0.681 & 0.715 & 0.735 & 0.747 & 0.753 & 0.757 & 0.760 & 0.763 & 0.767 & 0.728 \\
TP-Net \textcolor{lightgray}{\scriptsize{[TPAMI'23]}}~\citep{tan2023temporal} &0.552&0.663&0.713&0.738&0.757&0.765&\textbf{0.774}&\textbf{0.779}&\textbf{0.784}&\textbf{0.788}&0.732 \\
MA-Net \textcolor{lightgray}{\scriptsize{[WACV'23]}}~\citep{MA-net}&0.375&0.502&0.569&0.624&0.658&0.677&0.695&0.703&0.711&0.717&0.623\\
BasicGEBD \textcolor{lightgray}{\scriptsize{[ACM MM'24]}}~\citep{zheng2024rethinking}&0.600&0.666&-&-&-&0.731&-&-&-&0.748&0.710\\
\midrule
Ours &\textbf{0.626} &\textbf{0.702} &\textbf{0.731} &\textbf{0.753} &\textbf{0.762} &\textbf{0.768} &0.771 &0.774 &0.778 &0.780& \textbf{0.745}\\
\specialrule{1.5pt}{0pt}{0pt}
\end{tabular}
}
\caption{F1 results on TAPOS validation split with Rel.Dis. threshold set from 0.05 to 0.5 with 0.05 interval.}
\label{tab:tapos_val}
\end{table*}

\begin{table}[!t]
    \centering
    \setlength{\tabcolsep}{4.5pt}
    \scalebox{1.0}{
    \begin{tabular}{l|ccc|c}
    \rowcolor{mygray}
    \specialrule{1.5pt}{0pt}{0pt}
        Model &ClipShots &BBC &RAI & speed (ms)\\
    \hline
         TransNet &73.5 &92.9 &94.3 & - \\
         ST-CNN &75.9 &92.6 &93.9& -  \\
         DSM &76.1 &89.3 &92.8 &1.3  \\
         TransNet V2 &77.9 &96.2 &93.9&1.4   \\
         \hline
         Ours&\textbf{79.5} &\textbf{98.4} &\textbf{96.2} & \textbf{0.9} \\
    \specialrule{1.5pt}{0pt}{0pt}
    \end{tabular}
    }
    \caption{F1 scores on ClipShots, BBC and RAI datasets.}
    \label{tab:shot_results}
    \vspace{-0.2cm}
\end{table}

\begin{table*}[!t]
    \centering
    \renewcommand{\arraystretch}{1.2}
    \setlength{\tabcolsep}{6pt}
    \scalebox{1.0}{
    \begin{tabular}{l|cc|cc|cc|cc}
    \rowcolor{mygray}
    \specialrule{1.5pt}{0pt}{0pt}
         Representation  & 0.05 &$\Delta$  &0.25 &$\Delta$& 0.5&$\Delta$ & avg &$\Delta$ \\
    \hline
         1D CNN&0.609&-0.175 &0.838&-0.058 &0.864&-0.044 &0.810&-0.073  \\
         Swin~\citep{DBLP:conf/iccv/swin-transformer}&0.703&-0.081 &0.870&-0.026 &0.891&-0.017 & 0.849&-0.034  \\
         SPoS&\textbf{0.784}&- &\textbf{0.896}&- &\textbf{0.911}&- &\textbf{0.883}&- \\
    \specialrule{1.5pt}{0pt}{0pt}
    \end{tabular}
    }
    \caption{Importance of structured partition of sequence (SPoS). $\Delta$ columns show the differences with our SPoS. This verifies that SPoS is crucial for boundary detection.}
    \label{tab:ablation_window_representation}
\end{table*}

\begin{table*}[!t]
\centering
\subfloat[
Effect of adjacent window size $K$.
\label{tab:ablation_k}
]{
\centering
\begin{minipage}{0.32\linewidth}{
\begin{center}
\setlength{\tabcolsep}{4pt}
\begin{tabular}{c|cccc}
\rowcolor{mygray}
\specialrule{1.5pt}{0pt}{0pt}
    $K$  & 0.05 &0.25 & 0.5 & avg \\
\hline
     $4$&0.763 &0.881 &0.898 & 0.868  \\
     $6$&0.773 &0.890 &0.906 & 0.876 \\
     $8$&\baseline{\textbf{0.784}} &\baseline{\textbf{0.896}} &\baseline{\textbf{0.911}}&\baseline{\textbf{0.883}}  \\
     $10$&0.783 &0.895 &{\textbf{0.911}} &0.882  \\
     $12$&0.782 &{\textbf{0.896}} &{0.910} &{0.880}  \\
\specialrule{1.5pt}{0pt}{0pt}
\end{tabular}
\end{center}}\end{minipage}
}
\subfloat[
Effect of model width $C$.
\label{tab:ablation_channel}
]{
\begin{minipage}{0.32\linewidth}{\begin{center}
\setlength{\tabcolsep}{4pt}
\renewcommand{\arraystretch}{1.16}
\begin{tabular}{c|cccc}
\rowcolor{mygray}
\specialrule{1.5pt}{0pt}{0pt}
     $C$  & 0.05 &0.25 & 0.5 & avg \\
\hline
     $128$&0.781 &{0.894} &{0.910} &{0.881}  \\
     $256$&\baseline{\textbf{0.784}} &\baseline{\textbf{0.896}} &\baseline{\textbf{0.911}} & \baseline{\textbf{0.883}} \\
     $512$&0.782 &0.895 &\textbf{0.911} & 0.882  \\
     $768$&0.779  &0.883 &0.908 &0.879  \\
\specialrule{1.5pt}{0pt}{0pt}
\end{tabular}
\end{center}}\end{minipage}
}
\subfloat[
Effect of number of groups $G$.
\label{tab:ablation_group}
]{
\begin{minipage}{0.32\linewidth}{\begin{center}
\setlength{\tabcolsep}{4pt}
\renewcommand{\arraystretch}{1.1}
\begin{tabular}{c|cccc}
\rowcolor{mygray}
\specialrule{1.5pt}{0pt}{0pt}
     $G$  & 0.05 &0.25 & 0.5 & avg \\
\hline
     $1$&0.760 &0.871 &0.886 &0.860  \\
     $2$&0.772 &0.892 &0.908 & 0.878 \\
     $4$&\baseline{\textbf{0.784}} &\baseline{\textbf{0.896}} &\baseline{\textbf{0.911}} &\baseline{\textbf{0.883}}  \\
     $8$&0.781 &0.893 &0.910 &0.881  \\
\specialrule{1.5pt}{0pt}{0pt}
\end{tabular}
\end{center}}\end{minipage}
}
\caption{Our Structured Context Learning method ablation experiments on Kinetics-GEBD validation dataset. We report F1 scores for Rel.Dis. thresholds of 0.05, 0.25, and 0.5. The ``avg'' column represents the average F1 score across all Rel.Dis. thresholds ranging from 0.05 to 0.5, with an interval of 0.05. Default settings are marked in \colorbox{baselinecolor}{gray}.}
\label{tab:ablations}
\vspace{-0.4cm}
\end{table*}

\subsection{Ablations}

In the ablation analysis, we investigate the impact of each component of our method and the loss on the final performance. To conduct this study, we performed experiments on the Kinetics-GEBD dataset using ResNet-50 as the backbone. Due to space constraints, we only report F1 scores for Rel.Dis. thresholds of 0.05, 0.25, and 0.5. The column ``avg'' represents the average F1 score in all Rel.Dis. thresholds range from 0.05 to 0.5, with an interval of 0.05. The results are in Table~\ref{tab:ablations}. Default settings are marked in \colorbox{baselinecolor}{gray}.

\noindent{\textbf{Importance of SPoS.}}
The structured sequence partition provides a shared local temporal context for each frame to predict event boundaries. To verify its effectiveness, we attempted to completely remove it and used the 1D convolutional neural network and the shifted window (Swin) representation \citep{DBLP:conf/iccv/swin-transformer} as replacements. the results can be found in Table \ref{tab:ablation_window_representation}. From the table, we observed a significant performance drop after replacing SPoS. It can be interpreted that 1D CNNs only enlarge the receptive field of each candidate frame, and this impact is actually distributed as a Gaussian \citep{DBLP:conf/nips/LuoLUZ16}. This is not optimal for event boundary detection, since nearby frames may have equal importance. As for Swin, it is designed to relieve the Transformer's global self-attention computation burden by leveraging non-overlapping shifted windows. And each frame can attend to very distant frames after several Swin Transformer Block stacks. We think this is \texttt{not aligned} with the GEBD task, as adjacent frames are more important, while distant frames may cross multiple different boundaries and thus disturb the convergence. This also verifies that a structured representation is crucial for boundary detection.

\noindent{\textbf{Adjacent window size $K$.}} 
Adjacent window size $K$ defines how far the subsequent module can capture context information in the temporal domain. A smaller $K$ may not be able to capture enough necessary contextual information for a boundary, while a larger $K$ introduces noisy information when crossing two or more different boundaries. Table \ref{tab:ablation_k} presents the F1 scores obtained by varying $K$. The results suggest that different kinds of boundaries may prefer different window sizes, intuitively, as event boundaries in a video may span a different number of frames. Although adapting the $K$ size using a more sophisticated mechanism could further improve performance, we choose a fixed-length window in all our experiments for simplicity, leaving this for future work. We select $K=8$ as the adjacent window size, as the performance gain decreases as $K$ increases.

\noindent{\textbf{Effect of model width.}}
In Table \ref{tab:ablation_channel}, we analyze the effect of model width (\ie, the number of channels) on the performance of our proposed method. We experiment with five different channel numbers: $128$, $256$, $512$, and $768$. The results show that the model with $C=256$ achieves the best speed and accuracy trade-off, and we use it as the default channel number in our experiments.

\noindent{\textbf{Number of groups.}}
We evaluate the importance of group similarity by changing the number of groups $G$, results are shown in Table \ref{tab:ablation_group}. We observe steady performance improvements when increasing $G$ and saturation when $G=4$. This result shows the effectiveness of grouping channels when computing similarity.

\subsection{Qualitative Analysis}

\begin{figure*}[!t]
    \centering
    \includegraphics[width=0.85\textwidth]{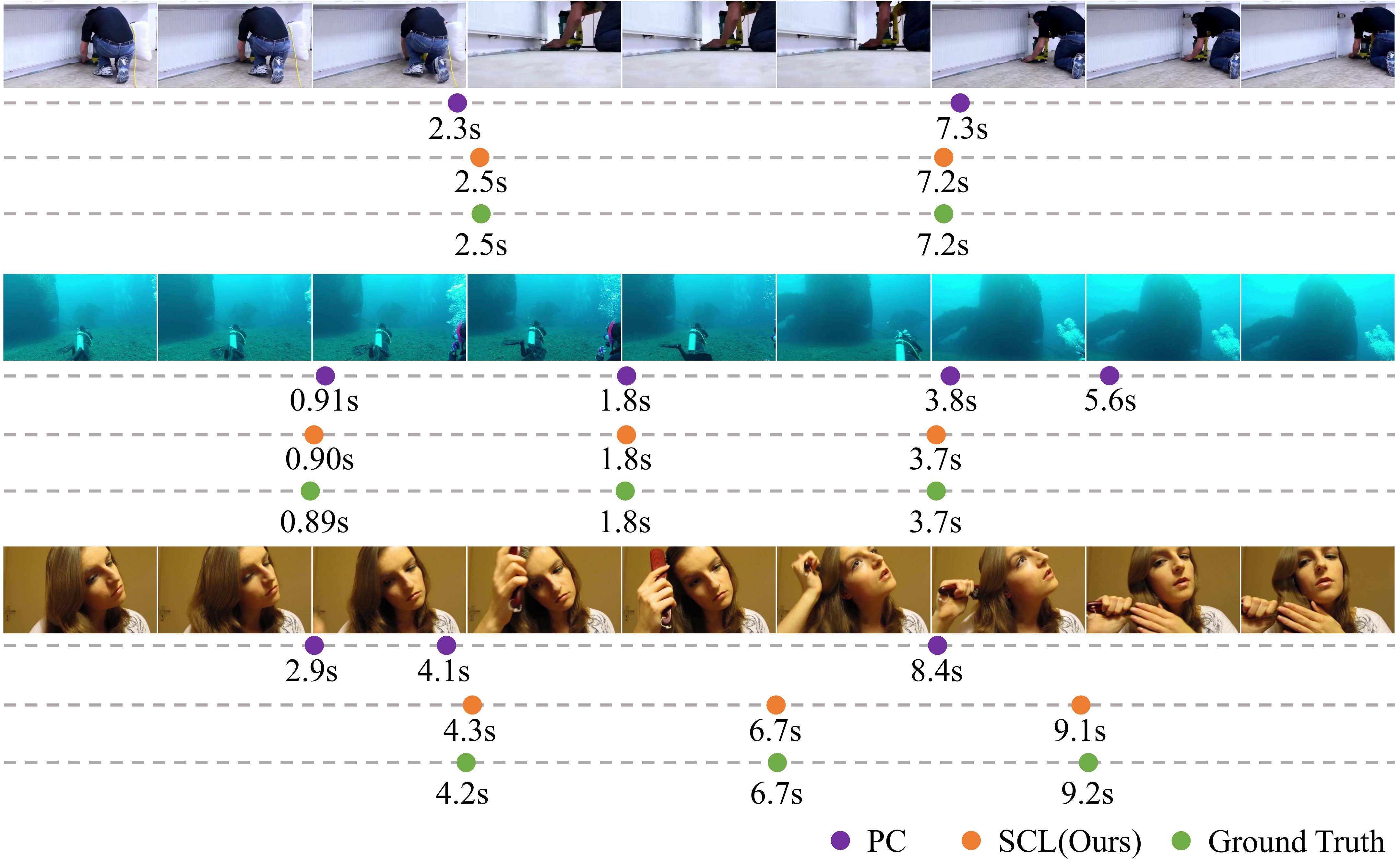}
    \caption{Example qualitative results on Kinetics-GEBD validation split. Best view in color.}
    \label{fig:vis}
    \vspace{-3mm}
\end{figure*}

The example qualitative results on Kinetics-GEBD are shown in Figure~\ref{fig:vis}. Compared with PC, our Structured Context Learning method can generate more accurate boundaries that are consistent with the ground truth, which further demonstrates the effectiveness of our method from a qualitative perspective. 

\section{Conclusions}
In this work, we propose a fully end-to-end method for Generic Event Boundary Detection (GEBD) by introducing Structured Context Learning with a designed Structured Partition of Sequence (SPoS) mechanism. The proposed method is flexible in choosing different temporal models and enables feature sharing by design. Experimental results on the challenging datasets Kinetics - GEBD and TAPOS demonstrate the superiority of the proposed method compared to existing state-of-the-art methods. Furthermore, our method also achieves promising results on shot transition detection datasets. Ablation studies are conducted to analyze the effectiveness of different components in the proposed method. Overall, the proposed method is highly efficient and effective, and we hope it can inspire future work in the field of GEBD.

\clearpage
{
    \small
    \bibliographystyle{ieeenat_fullname}
    \bibliography{main}

@String(CVPR= {IEEE Conf. Comput. Vis. Pattern Recog.})

@String(ICCV= {Int. Conf. Comput. Vis.})

@String(ECCV= {Eur. Conf. Comput. Vis.})

@String(NIPS= {Adv. Neural Inform. Process. Syst.})

@String(ACCV  = {ACCV})

@String(CVPR  = {CVPR})

@String(ICCV  = {ICCV})

@String(ECCV  = {ECCV})

@String(NIPS  = {NeurIPS})

@article{DBLP:journals/corr/GEBD,
  author    = {Mike Zheng Shou and
               Deepti Ghadiyaram and
               Weiyao Wang and
               Matt Feiszli},
  title     = {Generic Event Boundary Detection: {A} Benchmark for Event Segmentation},
  journal   = {CoRR},
  volume    = {abs/2101.10511},
  year      = {2021}
}

@inproceedings{DBLP:conf/cvpr/TAPOS,
  author    = {Dian Shao and
               Yue Zhao and
               Bo Dai and
               Dahua Lin},
  title     = {Intra- and Inter-Action Understanding via Temporal Action Parsing},
  booktitle = {CVPR},
  pages     = {727--736},
  publisher = {Computer Vision Foundation / {IEEE}},
  year      = {2020}
}

@inproceedings{DBLP:conf/cvpr/DMC-Net,
  author    = {Jiaqi Tang and
               Zhaoyang Liu and
               Chen Qian and
               Wayne Wu and
               Limin Wang},
  title     = {Progressive Attention on Multi-Level Dense Difference Maps for Generic
               Event Boundary Detection},
  booktitle = {{CVPR}},
  pages     = {3345--3354},
  publisher = {{IEEE}},
  year      = {2022}
}

@inproceedings{zhao2017temporal,
  title={Temporal action detection with structured segment networks},
  author={Zhao, Yue and Xiong, Yuanjun and Wang, Limin and Wu, Zhirong and Tang, Xiaoou and Lin, Dahua},
  booktitle={Proceedings of the IEEE International Conference on Computer Vision},
  pages={2914--2923},
  year={2017}
}

@inproceedings{DBLP:conf/nips/att_is_all_you_need,
  author    = {Ashish Vaswani and
               Noam Shazeer and
               Niki Parmar and
               Jakob Uszkoreit and
               Llion Jones and
               Aidan N. Gomez and
               Lukasz Kaiser and
               Illia Polosukhin},
  title     = {Attention is All you Need},
  booktitle = {NIPS},
  pages     = {5998--6008},
  year      = {2017}
}

@inproceedings{DBLP:conf/cvpr/FCN,
  author    = {Jonathan Long and
               Evan Shelhamer and
               Trevor Darrell},
  title     = {Fully convolutional networks for semantic segmentation},
  booktitle = {CVPR},
  pages     = {3431--3440},
  publisher = {{IEEE} Computer Society},
  year      = {2015}
}

@inproceedings{DBLP:conf/cvpr/resnet,
  author    = {Kaiming He and
               Xiangyu Zhang and
               Shaoqing Ren and
               Jian Sun},
  title     = {Deep Residual Learning for Image Recognition},
  booktitle = {CVPR},
  pages     = {770--778},
  publisher = {{IEEE} Computer Society},
  year      = {2016}
}

@article{DBLP:journals/corr/gebd-contrastive-learning,
  author    = {Hyolim Kang and
               Jinwoo Kim and
               Kyungmin Kim and
               Taehyun Kim and
               Seon Joo Kim},
  title     = {Winning the CVPR'2021 Kinetics-GEBD Challenge: Contrastive Learning
               Approach},
  journal   = {CoRR},
  volume    = {abs/2106.11549},
  year      = {2021}
}

@article{DBLP:journals/corr/gebd-UBoCo,
  author    = {Hyolim Kang and
               Jinwoo Kim and
               Taehyun Kim and
               Seon Joo Kim},
  title     = {UBoCo : Unsupervised Boundary Contrastive Learning for Generic Event
               Boundary Detection},
  journal   = {CoRR},
  volume    = {abs/2111.14799},
  year      = {2021}
}

@article{DBLP:journals/natmi/shortcut-learning,
  author    = {Robert Geirhos and
               J{\"{o}}rn{-}Henrik Jacobsen and
               Claudio Michaelis and
               Richard S. Zemel and
               Wieland Brendel and
               Matthias Bethge and
               Felix A. Wichmann},
  title     = {Shortcut learning in deep neural networks},
  journal   = {Nat. Mach. Intell.},
  volume    = {2},
  number    = {11},
  pages     = {665--673},
  year      = {2020}
}

@inproceedings{DBLP:conf/nips/alexnet,
  author    = {Alex Krizhevsky and
               Ilya Sutskever and
               Geoffrey E. Hinton},
  title     = {ImageNet Classification with Deep Convolutional Neural Networks},
  booktitle = {NIPS},
  pages     = {1106--1114},
  year      = {2012}
}

@inproceedings{DBLP:conf/cvpr/ResNeXt,
  author    = {Saining Xie and
               Ross B. Girshick and
               Piotr Doll{\'{a}}r and
               Zhuowen Tu and
               Kaiming He},
  title     = {Aggregated Residual Transformations for Deep Neural Networks},
  booktitle = {CVPR},
  pages     = {5987--5995},
  publisher = {{IEEE} Computer Society},
  year      = {2017}
}

@inproceedings{DBLP:conf/eccv/GN,
  author    = {Yuxin Wu and
               Kaiming He},
  title     = {Group Normalization},
  booktitle = {ECCV},
  volume    = {11217},
  pages     = {3--19},
  year      = {2018}
}

@inproceedings{DBLP:conf/cvpr/imagenet,
  author    = {Jia Deng and
               Wei Dong and
               Richard Socher and
               Li{-}Jia Li and
               Kai Li and
               Li Fei{-}Fei},
  title     = {ImageNet: {A} large-scale hierarchical image database},
  booktitle = {CVPR},
  pages     = {248--255},
  publisher = {{IEEE} Computer Society},
  year      = {2009}
}

@inproceedings{DBLP:conf/iccv/BMN,
  author    = {Tianwei Lin and
               Xiao Liu and
               Xin Li and
               Errui Ding and
               Shilei Wen},
  title     = {{BMN:} Boundary-Matching Network for Temporal Action Proposal Generation},
  booktitle = {ICCV},
  pages     = {3888--3897},
  publisher = {{IEEE}},
  year      = {2019}
}

@inproceedings{DBLP:conf/eccv/TCN,
  author    = {Colin Lea and
               Austin Reiter and
               Ren{\'{e}} Vidal and
               Gregory D. Hager},
  title     = {Segmental Spatiotemporal CNNs for Fine-Grained Action Segmentation},
  booktitle = {ECCV},
  volume    = {9907},
  pages     = {36--52},
  year      = {2016}
}

@inproceedings{caba2017scc,
  title={Scc: Semantic context cascade for efficient action detection},
  author={Caba Heilbron, Fabian and Barrios, Wayner and Escorcia, Victor and Ghanem, Bernard},
  booktitle={CVPR},
  pages={1454--1463},
  year={2017}
}

@InProceedings{Chao_2018_CVPR,
author = {Chao, Yu-Wei and Vijayanarasimhan, Sudheendra and Seybold, Bryan and Ross, David A. and Deng, Jia and Sukthankar, Rahul},
title = {Rethinking the Faster R-CNN Architecture for Temporal Action Localization},
booktitle = {CVPR},
month = {June},
year = {2018}
}

@inproceedings{alwassel2018action,
  title={Action search: Spotting actions in videos and its application to temporal action localization},
  author={Alwassel, Humam and Heilbron, Fabian Caba and Ghanem, Bernard},
  booktitle={ECCV},
  pages={251--266},
  year={2018}
}

@inproceedings{lin2017single,
  title={Single shot temporal action detection},
  author={Lin, Tianwei and Zhao, Xu and Shou, Zheng},
  booktitle={Proceedings of the 25th ACM international conference on Multimedia},
  pages={988--996},
  year={2017}
}

@inproceedings{long2019gaussian,
  title={Gaussian temporal awareness networks for action localization},
  author={Long, Fuchen and Yao, Ting and Qiu, Zhaofan and Tian, Xinmei and Luo, Jiebo and Mei, Tao},
  booktitle={CVPR},
  pages={344--353},
  year={2019}
}

@inproceedings{zhao2020bottom,
  title={Bottom-up temporal action localization with mutual regularization},
  author={Zhao, Peisen and Xie, Lingxi and Ju, Chen and Zhang, Ya and Wang, Yanfeng and Tian, Qi},
  booktitle={ECCV},
  pages={539--555},
  year={2020},
  organization={Springer}
}

@InProceedings{Yuan_2017_CVPR,
author = {Yuan, Zehuan and Stroud, Jonathan C. and Lu, Tong and Deng, Jia},
title = {Temporal Action Localization by Structured Maximal Sums},
booktitle = {CVPR},
month = {July},
year = {2017}
}

@inproceedings{DBLP:conf/iccv/swin-transformer,
   author    = {Ze Liu and
               Yutong Lin and
               Yue Cao and
               Han Hu and
               Yixuan Wei and
               Zheng Zhang and
               Stephen Lin and
               Baining Guo},
    title     = {Swin Transformer: Hierarchical Vision Transformer Using Shifted Windows},
    booktitle = {ICCV},
    month     = {October},
    pages     = {10012-10022},
    year      = {2021}
}

@article{DBLP:journals/corr/Longformer,
  author    = {Iz Beltagy and
               Matthew E. Peters and
               Arman Cohan},
  title     = {Longformer: The Long-Document Transformer},
  journal   = {CoRR},
  volume    = {abs/2004.05150},
  year      = {2020}
}

@inproceedings{DBLP:conf/cvpr/TCANet,
  author    = {Zhiwu Qing and
               Haisheng Su and
               Weihao Gan and
               Dongliang Wang and
               Wei Wu and
               Xiang Wang and
               Yu Qiao and
               Junjie Yan and
               Changxin Gao and
               Nong Sang},
  title     = {Temporal Context Aggregation Network for Temporal Action Proposal
               Refinement},
  booktitle = {{CVPR}},
  pages     = {485--494},
  publisher = {Computer Vision Foundation / {IEEE}},
  year      = {2021}
}

@article{DBLP:journals/corr/FaceTransformer,
  author    = {Yaoyao Zhong and
               Weihong Deng},
  title     = {Face Transformer for Recognition},
  journal   = {CoRR},
  volume    = {abs/2103.14803},
  year      = {2021}
}

@inproceedings{DBLP:conf/iccv/T2TViT,
  author    = {Li Yuan and
               Yunpeng Chen and
               Tao Wang and
               Weihao Yu and
               Yujun Shi and
               Zihang Jiang and
               Francis E. H. Tay and
               Jiashi Feng and
               Shuicheng Yan},
  title     = {Tokens-to-Token ViT: Training Vision Transformers from Scratch on
               ImageNet},
  booktitle = {{ICCV}},
  pages     = {538--547},
  publisher = {{IEEE}},
  year      = {2021}
}

@inproceedings{DBLP:conf/nips/LuoLUZ16,
  author    = {Wenjie Luo and
               Yujia Li and
               Raquel Urtasun and
               Richard S. Zemel},
  title     = {Understanding the Effective Receptive Field in Deep Convolutional
               Neural Networks},
  booktitle = {NIPS},
  pages     = {4898--4906},
  year      = {2016}
}

@inproceedings{DBLP:conf/icml/3D-Convolutional,
  author    = {Shuiwang Ji and
               Wei Xu and
               Ming Yang and
               Kai Yu},
  title     = {3D Convolutional Neural Networks for Human Action Recognition},
  booktitle = {ICML},
  pages     = {495--502},
  publisher = {Omnipress},
  year      = {2010}
}

@inproceedings{DBLP:conf/iccv/Learning-Spatiotemporal,
  author    = {Du Tran and
               Lubomir D. Bourdev and
               Rob Fergus and
               Lorenzo Torresani and
               Manohar Paluri},
  title     = {Learning Spatiotemporal Features with 3D Convolutional Networks},
  booktitle = {ICCV},
  pages     = {4489--4497},
  publisher = {{IEEE} Computer Society},
  year      = {2015}
}

@inproceedings{DBLP:conf/iccv/SlowFast,
  author    = {Christoph Feichtenhofer and
               Haoqi Fan and
               Jitendra Malik and
               Kaiming He},
  title     = {SlowFast Networks for Video Recognition},
  booktitle = {ICCV},
  pages     = {6201--6210},
  publisher = {{IEEE}},
  year      = {2019}
}

@inproceedings{DBLP:conf/cvpr/Two-Stream,
  author    = {Christoph Feichtenhofer and
               Axel Pinz and
               Andrew Zisserman},
  title     = {Convolutional Two-Stream Network Fusion for Video Action Recognition},
  booktitle = {CVPR},
  pages     = {1933--1941},
  publisher = {{IEEE} Computer Society},
  year      = {2016}
}

@article{DBLP:journals/corr/abs-2102-00719,
  author    = {Daniel Neimark and
               Omri Bar and
               Maya Zohar and
               Dotan Asselmann},
  title     = {Video Transformer Network},
  journal   = {CoRR},
  volume    = {abs/2102.00719},
  year      = {2021}
}

@article{DBLP:journals/corr/abs-2106-13230,
  author    = {Ze Liu and
               Jia Ning and
               Yue Cao and
               Yixuan Wei and
               Zheng Zhang and
               Stephen Lin and
               Han Hu},
  title     = {Video Swin Transformer},
  journal   = {CoRR},
  volume    = {abs/2106.13230},
  year      = {2021}
}

@article{DBLP:journals/corr/Kinetics,
  author    = {Will Kay and
               Jo{\~{a}}o Carreira and
               Karen Simonyan and
               Brian Zhang and
               Chloe Hillier and
               Sudheendra Vijayanarasimhan and
               Fabio Viola and
               Tim Green and
               Trevor Back and
               Paul Natsev and
               Mustafa Suleyman and
               Andrew Zisserman},
  title     = {The Kinetics Human Action Video Dataset},
  journal   = {CoRR},
  volume    = {abs/1705.06950},
  year      = {2017}
}

@article{DBLP:journals/corr/UCF101,
  author    = {Khurram Soomro and
               Amir Roshan Zamir and
               Mubarak Shah},
  title     = {{UCF101:} {A} Dataset of 101 Human Actions Classes From Videos in
               The Wild},
  journal   = {CoRR},
  volume    = {abs/1212.0402},
  year      = {2012}
}

@inproceedings{DBLP:conf/iccv/HMDB,
  author    = {Hildegard Kuehne and
               Hueihan Jhuang and
               Est{\'{\i}}baliz Garrote and
               Tomaso A. Poggio and
               Thomas Serre},
  title     = {{HMDB:} {A} large video database for human motion recognition},
  booktitle = {ICCV},
  pages     = {2556--2563},
  publisher = {{IEEE} Computer Society},
  year      = {2011}
}

@inproceedings{DBLP:conf/cvpr/PerazziPMGGS16,
  author    = {Federico Perazzi and
               Jordi Pont{-}Tuset and
               Brian McWilliams and
               Luc Van Gool and
               Markus H. Gross and
               Alexander Sorkine{-}Hornung},
  title     = {A Benchmark Dataset and Evaluation Methodology for Video Object Segmentation},
  booktitle = {CVPR},
  pages     = {724--732},
  publisher = {{IEEE} Computer Society},
  year      = {2016}
}

@article{tversky2013event,
  title={Event perception},
  author={Tversky, Barbara and Zacks, Jeffrey M},
  journal={Oxford handbook of cognitive psychology},
  volume={1},
  number={2},
  pages={3},
  year={2013},
  publisher={Oxford Oxford}
}

@article{DBLP:journals/access/XiaZ20,
  author    = {Huifen Xia and
               Yongzhao Zhan},
  title     = {A Survey on Temporal Action Localization},
  journal   = {{IEEE} Access},
  volume    = {8},
  pages     = {70477--70487},
  year      = {2020}
}

@article{DBLP:journals/corr/abs-2107-00239,
  author    = {Dexiang Hong and
               Congcong Li and
               Longyin Wen and
               Xinyao Wang and
               Libo Zhang},
  title     = {Generic Event Boundary Detection Challenge at {CVPR} 2021 Technical
               Report: Cascaded Temporal Attention Network {(CASTANET)}},
  journal   = {CoRR},
  volume    = {abs/2107.00239},
  year      = {2021}
}

@article{DBLP:journals/corr/abs-2106-10090,
  author    = {Ayush K. Rai and
               Tarun Krishna and
               Julia Dietlmeier and
               Kevin McGuinness and
               Alan F. Smeaton and
               Noel E. O'Connor},
  title     = {Discerning Generic Event Boundaries in Long-Form Wild Videos},
  journal   = {CoRR},
  volume    = {abs/2106.10090},
  year      = {2021}
}

@inproceedings{DBLP:conf/accv/TangFKCZ18,
  author    = {Shitao Tang and
               Litong Feng and
               Zhanghui Kuang and
               Yimin Chen and
               Wei Zhang},
  title     = {Fast Video Shot Transition Localization with Deep Structured Models},
  booktitle = {ACCV},
  series    = {Lecture Notes in Computer Science},
  volume    = {11361},
  pages     = {577--592},
  publisher = {Springer},
  year      = {2018}
}

@article{DBLP:journals/neco/lstm,
  author    = {Sepp Hochreiter and
               J{\"{u}}rgen Schmidhuber},
  title     = {Long Short-Term Memory},
  journal   = {Neural Comput.},
  volume    = {9},
  number    = {8},
  pages     = {1735--1780},
  year      = {1997}
}

@article{DBLP:journals/corr/GRU,
  author    = {Junyoung Chung and
               {\c{C}}aglar G{\"{u}}l{\c{c}}ehre and
               KyungHyun Cho and
               Yoshua Bengio},
  title     = {Empirical Evaluation of Gated Recurrent Neural Networks on Sequence
               Modeling},
  journal   = {CoRR},
  volume    = {abs/1412.3555},
  year      = {2014}
}

@article{DBLP:journals/corr/abs-2008-04838,
  author    = {Tom{\'{a}}s Soucek and
               Jakub Lokoc},
  title     = {TransNet {V2:} An effective deep network architecture for fast shot
               transition detection},
  journal   = {CoRR},
  volume    = {abs/2008.04838},
  year      = {2020}
}

@inproceedings{DBLP:conf/mm/BBC_Dataset,
  author    = {Lorenzo Baraldi and
               Costantino Grana and
               Rita Cucchiara},
  title     = {A Deep Siamese Network for Scene Detection in Broadcast Videos},
  booktitle = {{ACM} Multimedia},
  pages     = {1199--1202},
  publisher = {{ACM}},
  year      = {2015}
}

@inproceedings{DBLP:conf/caip/RAI_Dataset,
  author    = {Lorenzo Baraldi and
               Costantino Grana and
               Rita Cucchiara},
  title     = {Shot and Scene Detection via Hierarchical Clustering for Re-using
               Broadcast Video},
  booktitle = {{CAIP} {(1)}},
  series    = {Lecture Notes in Computer Science},
  volume    = {9256},
  pages     = {801--811},
  publisher = {Springer},
  year      = {2015}
}

@inproceedings{DBLP:conf/eccv/ShuffleNetV2,
  author    = {Ningning Ma and
               Xiangyu Zhang and
               Hai{-}Tao Zheng and
               Jian Sun},
  title     = {ShuffleNet {V2:} Practical Guidelines for Efficient {CNN} Architecture
               Design},
  booktitle = {{ECCV} {(14)}},
  series    = {Lecture Notes in Computer Science},
  volume    = {11218},
  pages     = {122--138},
  publisher = {Springer},
  year      = {2018}
}

@inproceedings{MA-net,
  title={Motion Aware Self-Supervision for Generic Event Boundary Detection},
  author={Rai, Ayush K and Krishna, Tarun and Dietlmeier, Julia and McGuinness, Kevin and Smeaton, Alan F and O’Connor, Noel E},
  booktitle={Proceedings of the IEEE/CVF Winter Conference on Applications of Computer Vision},
  pages={2728--2739},
  year={2023}
}

@inproceedings{li2022end,
  title={End-to-end compressed video representation learning for generic event boundary detection},
  author={Li, Congcong and Wang, Xinyao and Wen, Longyin and Hong, Dexiang and Luo, Tiejian and Zhang, Libo},
  booktitle={Proceedings of the IEEE/CVF Conference on Computer Vision and Pattern Recognition},
  pages={13967--13976},
  year={2022}
}

@inproceedings{shou2021generic,
  title={Generic event boundary detection: A benchmark for event segmentation},
  author={Shou, Mike Zheng and Lei, Stan Weixian and Wang, Weiyao and Ghadiyaram, Deepti and Feiszli, Matt},
  booktitle={Proceedings of the IEEE/CVF International Conference on Computer Vision},
  pages={8075--8084},
  year={2021}
}

@article{tan2023temporal,
  title={Temporal Perceiver: A General Architecture for Arbitrary Boundary Detection},
  author={Tan, Jing and Wang, Yuhong and Wu, Gangshan and Wang, Limin},
  journal={IEEE Transactions on Pattern Analysis and Machine Intelligence},
  year={2023},
  publisher={IEEE}
}

@inproceedings{kang2022uboco,
  title={Uboco: Unsupervised boundary contrastive learning for generic event boundary detection},
  author={Kang, Hyolim and Kim, Jinwoo and Kim, Taehyun and Kim, Seon Joo},
  booktitle={Proceedings of the IEEE/CVF Conference on Computer Vision and Pattern Recognition},
  pages={20073--20082},
  year={2022}
}

@inproceedings{zheng2024rethinking,
  title={Rethinking the Architecture Design for Efficient Generic Event Boundary Detection},
  author={Zheng, Ziwei and Zhang, Zechuan and Wang, Yulin and Song, Shiji and Huang, Gao and Yang, Le},
  booktitle={Proceedings of the 32nd ACM International Conference on Multimedia},
  pages={1215--1224},
  year={2024}
}
}

\clearpage

\maketitlesupplementary

In this supplementary material, we provide additional ablation results to further validate the effectiveness of our method.

\section{Additional Ablation Experiments}

\noindent{\textbf{A. Effect of temporal model.}}
The design of the Structured Partition of Sequence (SPoS) allows for flexibility in choosing different temporal models to achieve a better speed-accuracy trade-off. In addition to Transformers, we found that Recurrent Neural Networks (RNNs) also achieve good performance. As shown in Table~\ref{tab:ablation_temporal_model}, LSTM and GRU achieve competitive results and run faster than Transformers. We infer that SPoS provides localized structured context information for boundary detection, which is critical for accurate predictions. The subsequent temporal model is responsible for semantic learning. As a result, different temporal models have only a minor influence on the final performance. This confirms the effectiveness of SPoS and provides more options for selecting temporal models.

\noindent{\textbf{B. Effect of loss function.}} The GEBD task can be interpreted as a binary classification task at the frame level, where the goal is to classify each frame as a boundary or non-boundary after capturing temporal context information. To train our model, we use binary cross-entropy (BCE) loss and mean squared error (MSE) loss, with the option of turning Gaussian smoothing on or off (as introduced in section 3.3). As shown in Table \ref{tab:ablation_loss}, we observe that Gaussian smoothing can improve performance in both settings, indicating its effectiveness. We attribute this improvement to two factors: 1) Consecutive frames often have similar feature representations in the latent space, leading to a tendency for consecutive frames to output similar responses. Hard labels violate this tendency and may lead to poor convergence. 2) The annotations for GEBD are inherently ambiguous, and Gaussian smoothing can prevent the network from becoming overconfident. For all our experiments, we use the ``BCE + Gaussian" setting.

\begin{table}[h]
\begin{minipage}{.5\textwidth}
\begin{subtable}{\textwidth}
\renewcommand{\arraystretch}{1.3}
\scalebox{0.85}{
\begin{tabular}{l|cccc|c}
\rowcolor{mygray}
\specialrule{1.5pt}{0pt}{0pt}
Temporal model  & 0.05 &0.25 & 0.5 & avg &speed(ms) \\ \hline
Transformer &\baseline{\textbf{0.784}} &\baseline{\textbf{0.896}} &\baseline{\textbf{0.911}} &\baseline{\textbf{0.883}} &\baseline{\textbf{1.9}} \\
LSTM &0.772 &0.893 &0.909 & 0.879&1.1  \\
GRU &0.773 &0.894 &0.909 & 0.880&1.0 \\
\specialrule{1.5pt}{0pt}{0pt}
\end{tabular}}
\caption{Effect of temporal model.}
\label{tab:ablation_temporal_model}
\end{subtable}
\end{minipage}

\begin{minipage}{0.5\textwidth}
\begin{subtable}{\textwidth}%
\renewcommand{\arraystretch}{1.3}
\scalebox{0.95}{
\begin{tabular}{ccc|cccc}
\rowcolor{mygray}
\specialrule{1.5pt}{0pt}{0pt}
     BCE & MSE &Gaussian  & 0.05 &0.25 & 0.5 & avg \\ \hline
    &\checkmark&  &0.762 &0.885 &0.901 &0.869  \\
    &\checkmark&\checkmark  &0.775 &0.894 &0.909 &0.880  \\
    \checkmark &&&0.773 &0.892 &0.907 &0.878  \\
    \checkmark &&\checkmark &\baseline{\textbf{0.784}} &\baseline{\textbf{0.896}} &\baseline{\textbf{0.911}} &\baseline{\textbf{0.883}}  \\
\specialrule{1.5pt}{0pt}{0pt}
\end{tabular}}
\caption{Effect of loss function.}
\label{tab:ablation_loss}
\end{subtable}
\end{minipage}

\begin{minipage}{0.5\textwidth}
\begin{subtable}{\textwidth}%
\renewcommand{\arraystretch}{1.3}
\scalebox{1.0}{
\begin{tabular}{l|cccc}
\rowcolor{mygray}
    \specialrule{1.5pt}{0pt}{0pt}
          Similarity Function  & 0.05 &0.25 & 0.5 & avg \\
         \hline
         Chebyshev & 0.773 &0.889 &0.906 &0.877  \\
         Manhattan & 0.781 &0.894 &0.908 &0.880  \\
         Euclidean & 0.783 &0.895 &0.910 &0.882 \\
         Cosine   &\baseline{\textbf{0.784}} &\baseline{\textbf{0.896} }&\baseline{\textbf{0.911}} &\baseline{\textbf{0.883}} \\
    \specialrule{1.5pt}{0pt}{0pt}
    \end{tabular}}
\caption{Effect of $\text{similarity-function}(\cdot, \cdot)$ in Equation (3)}
\label{tab:ablation_sim_func}
\end{subtable}
\end{minipage}
\caption{Our Structured Context Learning method ablation experiments on Kinetics-GEBD validation dataset. We report F1 scores for Rel.Dis. thresholds of 0.05, 0.25, and 0.5. The ``avg'' column represents the average F1 score across all Rel.Dis. thresholds ranging from 0.05 to 0.5, with an interval of 0.05. Default settings are marked in \colorbox{baselinecolor}{gray}.}
\label{tab:ablations}
\end{table}

\noindent{\textbf{C. Effect of similarity function.}} We investigate the impact of different distance metrics on our proposed method, which we refer to as similarity metrics since we use negative values to calculate them. Specifically, we evaluated four commonly used distance metrics, namely Cosine, Euclidean, Manhattan, and Chebyshev. It should be noted that each distance metric has its unique properties that can influence the performance of the model. For example, the Euclidean distance is sensitive to outliers, whereas the cosine distance is not. The Manhattan distance is more suitable for measuring the distance between two points in a grid-like structure. The experimental results, presented in Table \ref{tab:ablation_sim_func}, demonstrate that our method is effective across all four metrics. However, we chose the Cosine metric for our experiments due to its better performance compared to the other metrics.

\begin{table*}[htb]
\centering
\scalebox{1.0}{
\begin{tabular}{lcccccccccc|c}
\specialrule{1.5pt}{0pt}{0pt}
Rel.Dis. threshold& 0.05 & 0.1 & 0.15 & 0.2 & 0.25 & 0.3 & 0.35 & 0.4 & 0.45 & 0.5 & avg \\ 
\hline\hline
BMN~\citep{DBLP:conf/iccv/BMN}& 0.128& 0.141& 0.148& 0.152& 0.156& 0.159& 0.162 & 0.164 & 0.165& 0.167& 0.154 \\
BMN-StartEnd~\citep{DBLP:conf/iccv/BMN} & 0.396 &0.479 &0.509 &0.525 &0.534 &0.540 &0.544 &0.547 &0.549 &0.551 &0.517\\
TCN-TAPOS~\citep{DBLP:conf/eccv/TCN} & 0.518 &0.622 &0.665 &0.690 &0.706 &0.718 &0.727 &0.733 &0.738 &0.743 &0.686\\
TCN~\citep{DBLP:conf/eccv/TCN} & 0.461 &0.519 &0.538 &0.547 &0.553 &0.557 &0.559 &0.561 &0.563 &0.564 &0.542\\
PC~\citep{DBLP:journals/corr/GEBD}& 0.624 &0.753 &0.794 &0.816 &0.828 &0.836 &0.841 &0.844 &0.846 &0.849 &0.803\\
DDM-Net~\citep{DBLP:conf/cvpr/DMC-Net}&0.732& 0.812& 0.836& 0.849 &0.856 &0.860 &0.863 &0.865 &0.867& 0.869& 0.841\\
\hline
Ours& \textbf{0.754} & \textbf{0.831} & \textbf{0.850} & \textbf{0.862} &  \textbf{0.868} & \textbf{0.873} & \textbf{0.876} & \textbf{0.878}  & \textbf{0.880} & \textbf{0.883} &  \textbf{0.855} \\
\specialrule{1.5pt}{0pt}{0pt}
\end{tabular}
}
\caption{\textbf{Precision} on Kinetics-GEBD validation split with Rel.Dis. threshold set from 0.05 to 0.5 with 0.05 interval.}
\label{tab:gebd_val_prec}
\end{table*}

\begin{table*}[htb]
\centering
\scalebox{1.0}{
\begin{tabular}{lcccccccccc|c}
\specialrule{1.5pt}{0pt}{0pt}
Rel.Dis. threshold& 0.05 & 0.1 & 0.15 & 0.2 & 0.25 & 0.3 & 0.35 & 0.4 & 0.45 & 0.5 & avg \\ 
\hline\hline
BMN~\citep{DBLP:conf/iccv/BMN}& 0.338& 0.369 &0.385 &0.397 &0.407 &0.414 &0.420 &0.426 &0.430 &0.434 &0.402 \\
BMN-StartEnd~\citep{DBLP:conf/iccv/BMN} & 0.648 &0.766 &0.817 &0.846 &0.864 &0.876 &0.885 &0.892 &0.897 &0.900 &0.839\\
TCN-TAPOS~\citep{DBLP:conf/eccv/TCN} & 0.420 &0.508 &0.550 &0.576 &0.594 &0.609 &0.619 &0.627 &0.633 &0.639 &0.577\\
TCN~\citep{DBLP:conf/eccv/TCN} & 0.811 &0.894 &0.923 &0.938 &0.947 &0.952 &0.956 &0.959 &0.961 &0.963 &0.930\\
PC~\citep{DBLP:journals/corr/GEBD}& 0.626 &0.764 &0.814 &0.843 &0.859 &0.871 &0.879 &0.885 &0.889 &0.892 &0.832\\
DDM-Net~\citep{DBLP:conf/cvpr/DMC-Net}&0.800 &0.875 &0.899 &0.912 &0.920 &0.926 &0.930 &0.933 &0.935 &0.937 &0.907\\
\hline
Ours& \textbf{0.816} & \textbf{0.882} & \textbf{0.906} & \textbf{0.920} & \textbf{0.926} & \textbf{0.932} & \textbf{0.935} &  \textbf{0.937} & \textbf{0.940} & \textbf{0.941} & \textbf{0.913} \\
\specialrule{1.5pt}{0pt}{0pt}
\end{tabular}
}
\caption{\textbf{Recall} on Kinetics-GEBD validation split with Rel.Dis. threshold set from 0.05 to 0.5 with 0.05 interval.}
\label{tab:gebd_val_recall}
\end{table*}

\begin{table*}[!htbp]
\centering
\scalebox{1.0}{
\begin{tabular}{lcccccccccc|c}
\specialrule{1.5pt}{0pt}{0pt}
Rel.Dis. threshold& 0.05 & 0.1 & 0.15 & 0.2 & 0.25 & 0.3 & 0.35 & 0.4 & 0.45 & 0.5 & avg \\ 
\hline\hline
BMN~\citep{DBLP:conf/iccv/BMN}& 0.186  & 0.204  & 0.213  & 0.220  & 0.226& 0.230& 0.233& 0.237& 0.239& 0.241& 0.223\\
BMN-StartEnd~\citep{DBLP:conf/iccv/BMN} & 0.491& 0.589& 0.627& 0.648& 0.660& 0.668& 0.674& 0.678& 0.681& 0.683& 0.640\\
TCN-TAPOS~\citep{DBLP:conf/eccv/TCN} & 0.464& 0.560& 0.602& 0.628& 0.645& 0.659& 0.669& 0.676& 0.682& 0.687& 0.627\\
TCN~\citep{DBLP:conf/eccv/TCN} & 0.588& 0.657& 0.679& 0.691& 0.698& 0.703& 0.706& 0.708& 0.710& 0.712& 0.685\\
PC~\citep{DBLP:journals/corr/GEBD}& 0.625& 0.758&0.804&0.829&0.844&0.853&0.859&0.864&0.867&0.870&0.817\\
SBoCo-Res50~\citep{DBLP:journals/corr/gebd-UBoCo} &0.732&-&-&-&-&-&-&-&-&-&0.866 \\
DDM-Net~\citep{DBLP:conf/cvpr/DMC-Net}&0.764&0.843&0.866&0.880&0.887&0.892&0.895&0.898&0.900&0.902&0.873\\
\hline
Ours& \textbf{0.784} & \textbf{0.856} & \textbf{0.877} & \textbf{0.890} & \textbf{0.896} & \textbf{0.901} & \textbf{0.904} & \textbf{0.907} & \textbf{0.909} & \textbf{0.911} & \textbf{0.883} \\
\specialrule{1.5pt}{0pt}{0pt}
\end{tabular}
}
\caption{\textbf{F1} results on Kinetics-GEBD validation split with Rel.Dis. threshold set from 0.05 to 0.5 with 0.05 interval.}
\label{tab:gebd_val}
\end{table*}

\noindent{\textbf{D. Detailed results on Kinetics-GEBD.}} Table~\ref{tab:gebd_val_prec}, Table~\ref{tab:gebd_val_recall} and Table~\ref{tab:gebd_val} for Kinetics-GEBD respectively present the detailed results of precision, recall and f1 scores for various methods. It is noteworthy that some methods, like TCN, achieve high recall yet low precision because they make as many predictions as possible and recall many false positives. As a result, they do not achieve a superior F1 score.

\end{document}